\documentclass[fleqn,10pt]{wlscirep}
\usepackage[utf8]{inputenc}
\usepackage[T1]{fontenc}
\usepackage{graphicx}
\usepackage{subcaption}
\usepackage{booktabs}
\usepackage{siunitx}
\usepackage{dirtree}
\usepackage{booktabs}

\newlist{inlinelist}{enumerate*}{2}
\setlist[inlinelist,1,2]{label=(\roman*),align=left,leftmargin=8mm,labelsep*=0pt}

\DeclareSIUnit[quantity-product = \,]{\years}{\text{years}}

\usepackage{csquotes}

\title{UltraCortex: Submillimeter Ultra-High Field {9.4\,T} Brain MR Image Collection and Manual Cortical Segmentations}

\author[1,$\dag$,*]{Lucas~Mahler}
\author[2,1,$\dag$,*]{Julius~Steiglechner}
\author[3]{Benjamin~Bender}
\author[3,1]{Tobias~Lindig}
\author[1]{Dana~Ramadan}
\author[2,1]{Jonas~Bause}
\author[2,1]{Florian~Birk}
\author[1,2,4]{Rahel~Heule}
\author[1]{Edyta~Charyasz}
\author[2,1]{Michael~Erb}
\author[1]{Vinod~Jangir~Kumar}
\author[2,1]{Gisela~E~Hagberg}
\author[5]{Pascal~Martin}
\author[2,1]{Gabriele~Lohmann}
\author[2,1]{Klaus~Scheffler}

\affil[1]{Department High-field Magnetic Resonance, Max Planck Institute for Biological Cybernetics, T\"ubingen, Germany}
\affil[2]{Department of Biomedical Magnetic Resonance, University Hospital T\"ubingen, T\"ubingen, Germany}
\affil[3]{Department of Diagnostic and Interventional Neuroradiology, University Hospital T\"ubingen, T\"ubingen, Germany}
\affil[4]{Center for MR Research, University Children’s Hospital, Zurich, Switzerland}
\affil[5]{Department of Neurology and Epileptology, Hertie Institute for Clinical Brain Research, University of T\"ubingen, T\"ubingen, Germany}

\affil[*]{corresponding authors: Lucas Mahler (lucas.mahler@tue.mpg.de) and Julius Steiglechner (julius.steiglechner@tue.mpg.de)}
\affil[$\dag$]{these authors contributed equally to this work}

\begin{abstract}
    The \emph{UltraCortex} repository (\href{https://www.ultracortex.org/}{https://www.ultracortex.org}) houses magnetic resonance imaging data of the human brain obtained at an \text{ultra-high} field strength of \qty{9.4}{\tesla}.
  It contains \num{86}~structural MR images with spatial resolutions ranging from \qtyrange[range-units = single]{0.6}{0.8}{\mm}.
  Additionally, the repository includes segmentations of \num{12}~brains into gray and white matter compartments.
  These segmentations have been independently validated by two expert neuroradiologists, thus establishing them as a reliable gold standard.
  This resource provides researchers with access to high-quality brain imaging data and validated segmentations, facilitating neuroimaging studies and advancing our understanding of brain structure and function.
  Existing repositories do not accommodate field strengths beyond \qty{7}{\tesla}, nor do they offer validated segmentations, underscoring the significance of this new resource.
\end{abstract}
\begin{document}

\flushbottom
\maketitle

\thispagestyle{empty}

\phantomsection\addcontentsline{toc}{section}{Background \& Summary}
\section*{Background \& Summary}
Here we introduce the \emph{UltraCortex} repository (\href{https://www.ultracortex.org/}{https://www.ultracortex.org}), housing magnetic resonance imaging~(MRI) data of the human brain obtained at an \text{ultra-high} magnetic field strength of \qty{9.4}{\tesla}.
The repository comprises \num{86}~$T_1$-weighted structural MR~images of healthy human brains, acquired at spatial resolutions ranging from \qtyrange[range-units = single]{0.6}{0.8}{\mm} using \text{MP-RAGE} and MP2RAGE sequences.
Additionally, UltraCortex includes segmentations of \num{12}~brains into gray matter~(GM) and white matter~(WM) compartments.
These segmentations have undergone independent validation by two expert neuroradiologists, thus establishing them as a reliable gold standard.

While repositories containing MRI data at \qty{7}{\tesla}~exist~\cite{forstmann_multi-modal_2014, alkemade_amsterdam_2020, svanera_cerebrum-7t_2020}, none are currently available for field strengths beyond \qty{7}{\tesla}.
Moreover, validated segmentations of UHF~MRI data are presently lacking, but are urgently needed to support the development of new algorithms and tools for UHF~MRI data analysis, and enable the validation of existing methods.
It should be noted that we refer to field strengths $\ge \qty{7}{\tesla}$~as \emph{ultra-high field}~(UHF).

MRI has made a tremendous contribution to our understanding of the human brain.
However, \text{ultra-high} resolution brain imaging is constrained by technical challenges.
Image resolution and signal-to-noise ratio~(SNR) both increase with field strength, resulting in improved image quality at a UHF of \qty{9.4}{\tesla}\cite{pohmann_signal--noise_2016}.
This makes it easier to visualize structures with low inherent signal, such as small lesions or areas of low proton density.
The enhanced spatial resolution of UHF~MRI is beneficial for capturing fine anatomical details where conventional resolutions of \qtyrange[range-phrase = {\,--\,}, range-units = single]{1}{2}{\mm} are insufficient.
The improved tissue contrast of UHF~MRI allows for clearer classification of different tissue types, which is critical when studying various pathological conditions.
In addition, UHF~MRI offers enhanced sensitivity for functional MRI~(fMRI), allowing more precise localization of brain activity and more detailed spectroscopic imaging, which is useful for studying metabolites in conditions such as Alzheimer's disease and oncology.

The availability of large neuroimaging databases on open platforms such as OpenNeuro~(\href{https://openneuro.org/}{https://openneuro.org}), Image and Data Archive~(\href{https://ida.loni.usc.edu/}{https://ida.loni.usc.edu}), and Human Connectome Project\cite{van_essen_wu-minn_2013}~(\href{https://www.humanconnectome.org/}{https://humanconnectome.org}) has greatly advanced the field of neuroimaging.
These repositories have not only facilitated the development of imaging techniques, but have also provided a fertile ground for the development and application of new algorithms for the analysis of brain images.

However, the majority of existing data analysis algorithms were designed and optimized for MRI data acquired at lower field strengths, typically \qty{3.0}{\tesla} or \qty{1.5}{\tesla}.
In structural imaging, a spatial resolution of \qty{1}{\mm^3} has become the standard.
Consequently, when applied to UHF~MRI data, many algorithms use downsampling, resulting in a loss of the beneficial enhancements provided by UHF~MRI.
This mismatch highlights a bottleneck in UHF~MRI research: the underutilization of the detailed data that UHF~MRI can provide, constrained by the limitations of existing analysis tools.
The development and validation of new algorithms rely heavily on access to high-quality public databases, but there is currently no large public UHF~MRI database with field strengths of greater than \qty{7}{\tesla} and resolutions of smaller than \qty{1}{\mm}.
This becomes even more apparent when considering the availability of manually segmented MR images, the \emph{gold standard} for anatomical segmentation.
Only a small number of manually segmented images are available at $\le \qty{7}{\tesla}$\cite{klein_101_2012,kuijf_mr_2024,svanera_cerebrum-7t_2020}, and none are available at \qty{9.4}{\tesla}.
This limitation has significant implications for brain analysis, leading to inaccuracies in structural measures such as cortical volume, thickness, and depth, and thus potentially misleading results.
Furthermore, the application of these imaging techniques to fMRI analysis is also affected, as errors in the initial segmentation phase can cascade throughout the analysis pipeline.

\begin{figure}[ht]
    \centering
    \includegraphics[width=0.7\textwidth]{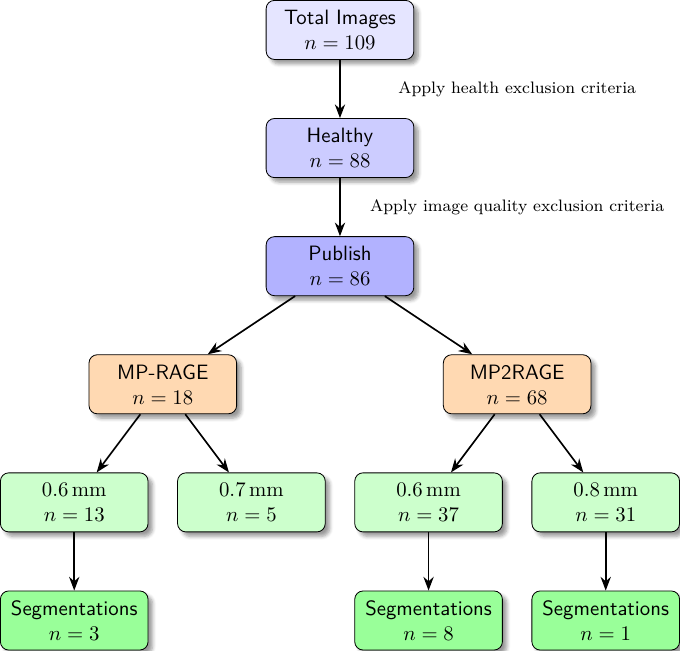}
    \caption{Dataset overview. The chart shows (i) from top to bottom how images were included or excluded for the subsequent steps of the pipeline, and (ii) the final distribution of the dataset across sequences, resolutions, and existing segmentations.}
    \label{fig:dataset_overview}
\end{figure}

With UltraCortex we aim to provide a foundation for the development and validation of new algorithms for UHF~MRI data analysis.
It can be used to develop new algorithms that are applicable across different field strengths and resolutions, as well as to validate and benchmark existing algorithms.
The data can also be used for image denoising, artifact removal, image quality enhancement and super-resolution.
Additionally, the data can be utilized for educational purposes, such as to illustrate the intricate details of brain structures.
The following sections describe the initial release of UltraCortex, including the data acquisition, processing and validation steps.
Figure~\ref{fig:dataset_overview} summarizes the dataset.

\phantomsection\addcontentsline{toc}{section}{Methods}
\section*{Methods}

\subsection*{Demographics}
Subsequent to the original measurements, we collected \num{86}~$T_1$-weighted structural brain MR images from \num{78}~healthy adult volunteers (\num{28}~females and \num{50}~males; age range: \qtyrange[range-phrase = {\,--\,}, range-units = single]{20.0}{53.0}{\years}; median age at acquisition: \qty{29.0}{\years}; interquartile range of age: \qty{7.8}{\years}).
\num{6}~subjects were measured in two independent sessions and \num{1}~subject was measured in three sessions.
The demographic data, including participant sex and age at the time of scanning, are presented in Table~\ref{tab:demographics}.
Some of the subject demographics and image data in this repository have been used in previous studies~\cite{charyasz_functional_2023}.

Participants were selected in accordance with the guidelines and ethics of the Max Planck Institute for Biological Cybernetics, T\"ubingen, Germany and the University Hospital, T\"ubingen, Germany.
Prior to enrollment in an MRI study, all subjects underwent a review of inclusion and exclusion criteria by a qualified physician. 
This included ensuring that, unless otherwise specified, there was no known history of neurological or psychiatric disease and that there were no contraindications to MRI scans.
Additionally, all participants were required to have normal or corrected-to-normal vision.
Prior to participation, written informed consent was obtained from all participants.
This consent included an agreement for open sharing of anonymized data.

Due to the more rigorous clinical criteria employed in this dataset, \num{21}~participants were excluded from an original collection of \num{109}~images from \num{101}~subjects.
Furhter, \num{2}~images were excluded due to a failed technical validation, resulting in the final dataset.

\subsection*{Data Acquisition}
\begin{figure}[ht]
  \includegraphics[width=\textwidth]{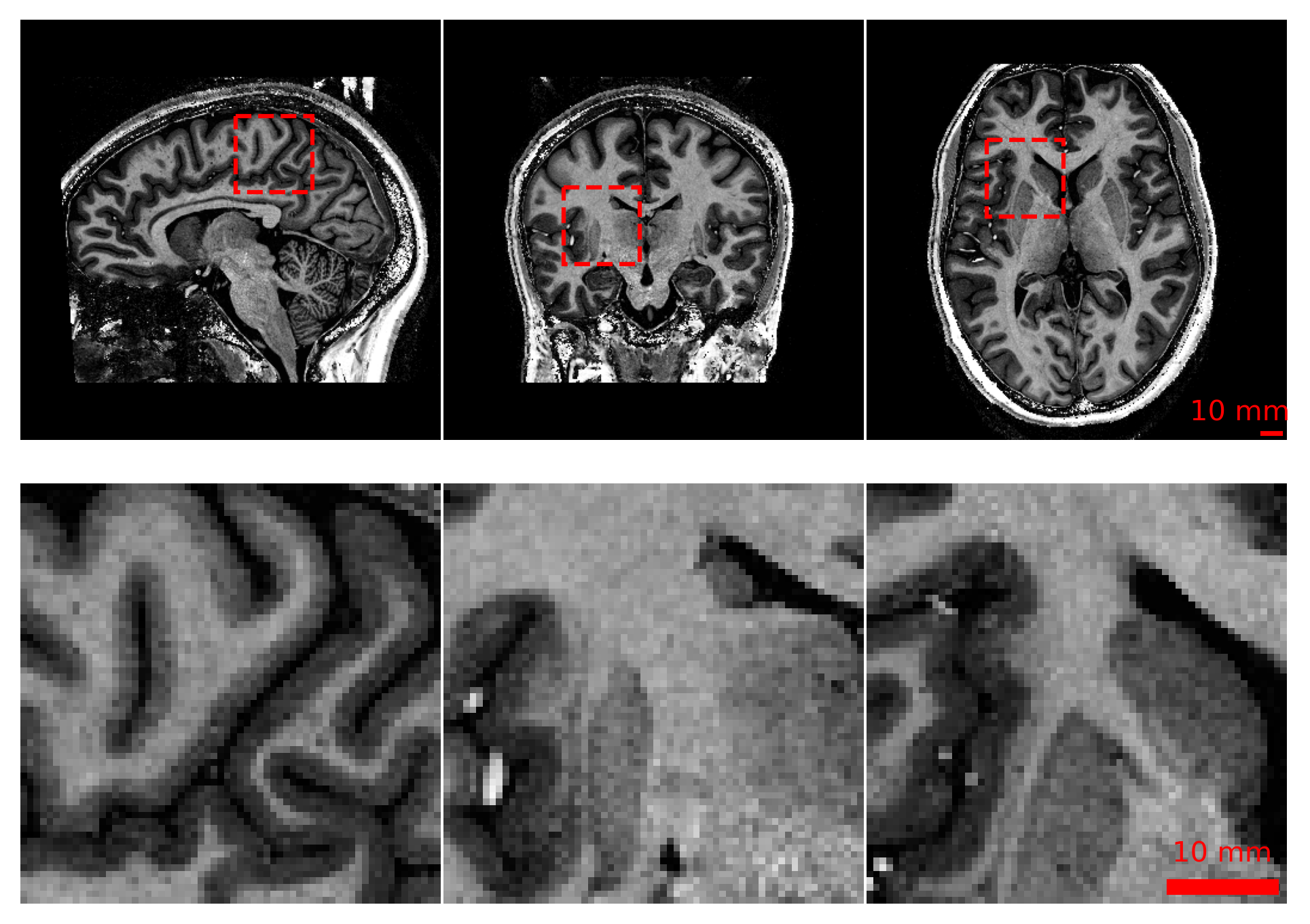}
    \caption{Example \qty{9.4}{\tesla} MRI slices from subject \texttt{sub-73} acquired with \qtyproduct[product-units = power]{0.6 x 0.6 x 0.6}{\mm}. The top row shows from left to right sagittal, coronal, and axial images and ROIs of the corresponding full resolution data in the bottom row.} 
    \label{fig:mri_slices}
\end{figure}

Data acquisition was conducted using a \qty{9.4}{\tesla} whole-body MRI scanner (Siemens Healthineers, Erlangen, Germany) at the Department for High-field Magnetic Resonance of the Max Planck Institute for Biological Cybernetics.
The scanner was equipped with a 16-channel dual-row transmit array, operating in CP+ (circularly polarized) mode, and paired with a 31-channel receive array\cite{shajan_16-channel_2014}.
After $B_0$ shimming, whole-brain $T_1$-weighted magnetization-prepared rapid acquisition gradient echo~(\text{MP-RAGE})\cite{mugler_iii_rapid_1991} or modified MP-RAGE~(MP2RAGE)\cite{marques_mp2rage_2010, hagberg_whole_2017} images were acquired.
Both sequences used a TR-FOCI pulse optimized for \qty{9.4}{\tesla} for magnetization inversion\cite{hurley_tailored_2010, hagberg_whole_2017}.
Nevertheless, despite the brief echo time used for the scans, signal dropouts due to strong $B_1$-field non-uniformity close to the ear canals were observed in all subjects.
It was determined that the data regarding the gradient nonlinearity profile would not be collected, indicating that the correction of gradient nonlinearity effects would not be conducted, as they play a relatively minor role within the field of view.
It is important to note, however, that quantitative analysis, such as volumetric measurements, may be affected by this.
The sequence parameters were empirically optimized for the contrast between GM and WM tissues.
An example of typical images is illustrated in Figure~\ref{fig:mri_slices}.
The parameter setups differ in sequence, number of acquired sagittal slices isotropic voxel size, acquisition matrix, inversion times $\text{TI1}$ and $\text{TI2}$, repetition time $\text{TR}$, echo time $\text{TE}$, flip angle $\text{FA}$, acceleration factor $\text{R}$ of GRAPPA\cite{griswold_generalized_2002} technology for parallel acquisition and reconstruction, slice partial Fourier factor $\text{PFF} = 6/8$, and optional CAIPI\cite{breuer_controlled_2005} shift in the GRAPPA technology.
An overview of the sequence configurations in use is presented in Table~\ref{tab:scanning_parameters}.
The comprehensive image-specific scanning parameters can be accessed via the data records.

\subsubsection*{MP-RAGE}
A total of \num{18}~whole-brain MP-RAGE images were acquired with five different sequence parameter setups and two different resolutions:
\begin{inlinelist}[mode=unboxed]
  \item \num{5}~images with $\text{voxel size} = \qtyproduct[product-units = power]{0.7 x 0.7 x 0.7}{\mm}$ and
  \item \num{13}~images with $\text{voxel size} = \qtyproduct[product-units = power]{0.6 x 0.6 x 0.6}{\mm}$.
\end{inlinelist}

\subsubsection*{MP2RAGE}
A total of \num{68}~whole-brain MP2RAGE images were acquired with six distinct sequence configurations and two different resolutions:
\begin{inlinelist}[mode=unboxed]
  \item \num{31}~images with $\text{voxel size} = \qtyproduct[product-units = power]{0.8 x 0.8 x 0.8}{\mm}$ and
  \item \num{37}~images with $\text{voxel size} = \qtyproduct[product-units = power]{0.6 x 0.6 x 0.6}{\mm}$.
\end{inlinelist}

\subsection*{Processing}
From each MP2RAGE scan, four distinct volumes were generated: INV1, INV2, UNI, and T1.
The UNI volume represents a composite of the images acquired at both inversion times (TI1 and TI2), whereas the T1 volume is a $T_1$ estimation map derived from the volumes at TI1 and TI2.
Background noise was removed from the T1 volume using a regularization term during contrast division when the images from both inversion times were combined~\cite{obrien_robust_2014}.
MP2RAGE images were reconstructed offline using customized routines that use the $B_1^+$ map when available to correct for the flip angle variation caused by the shorter wavelength at \qty{9.4}{\tesla}\cite{hagberg_whole_2017}.
The remaining $T_1$-weighted DICOM~(\href{https://www.dicomstandard.org/}{https://www.dicomstandard.org}) files were converted to \text{NIfTI-1}~(\href{https://nifti.nimh.nih.gov/nifti-1/}{https://nifti.nimh.nih.gov}) format using dcm2niix\cite{li_first_2016}.
The NIfTI files were then anonymized by first using SynthStrip\cite{hoopes_synthstrip_2022} to extract the brain mask and then using Quickshear\cite{schimke_quickshear_2011} to de-identify the images.
In addition, all images were stripped of any potentially revealing metadata in the NIfTI headers.

\subsection*{Segmentation}
We selected \num{12}~subjects whose ultra-high field \qty{9.4}{\tesla} MRI scans were of high quality for detailed manual segmentation.
The objective of the segmentation was to obtain high-precision masks of the cortical ribbon.
In accordance with FreeSurfer~\cite{fischl_whole_2002}, the labels shown in Table~\ref{tab:labels} were segmented.
This implies that subcortical structures, such as the basal ganglia, were added to the \enquote{inner} label, namely the cerebral white matter.
However, the hippocampi and amygdalae were added to the background according to the FreeSurfer ribbon file.
For cortical areas that exhibited signal dropouts, we opted to segment a topological closed cortex, leading to guessed labels in this region.

\begin{figure}[ht]
  \includegraphics[width=\textwidth]{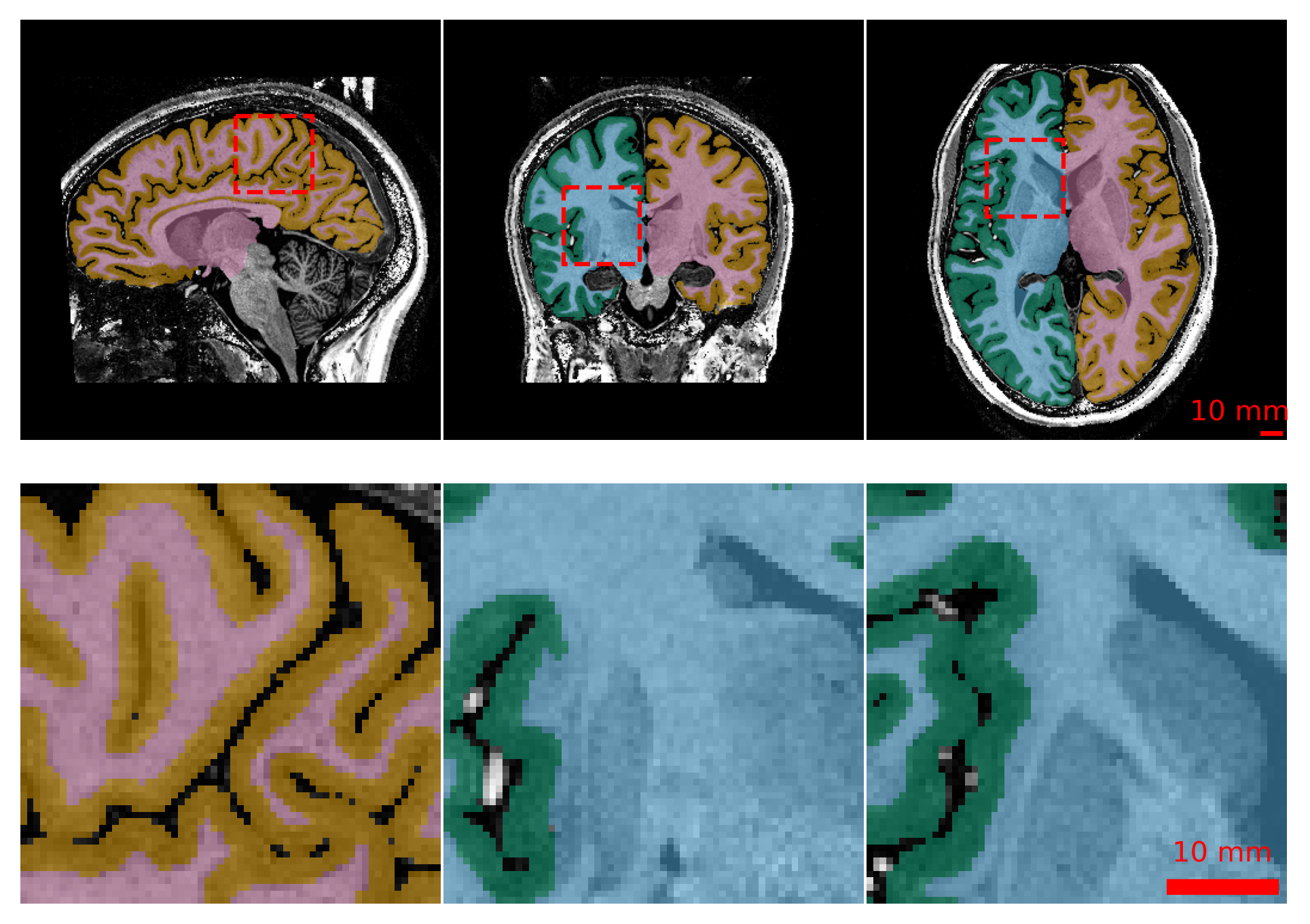}
    \caption{Results of the manually corrected cortical segmentation of Figure~\ref{fig:mri_slices}. Coloured labels are shown in overlay as described in Table~\ref{tab:labels}.}
    \label{fig:mri_slices_with_labels}
\end{figure}

\begin{figure}[ht]
  \includegraphics[width=\textwidth]{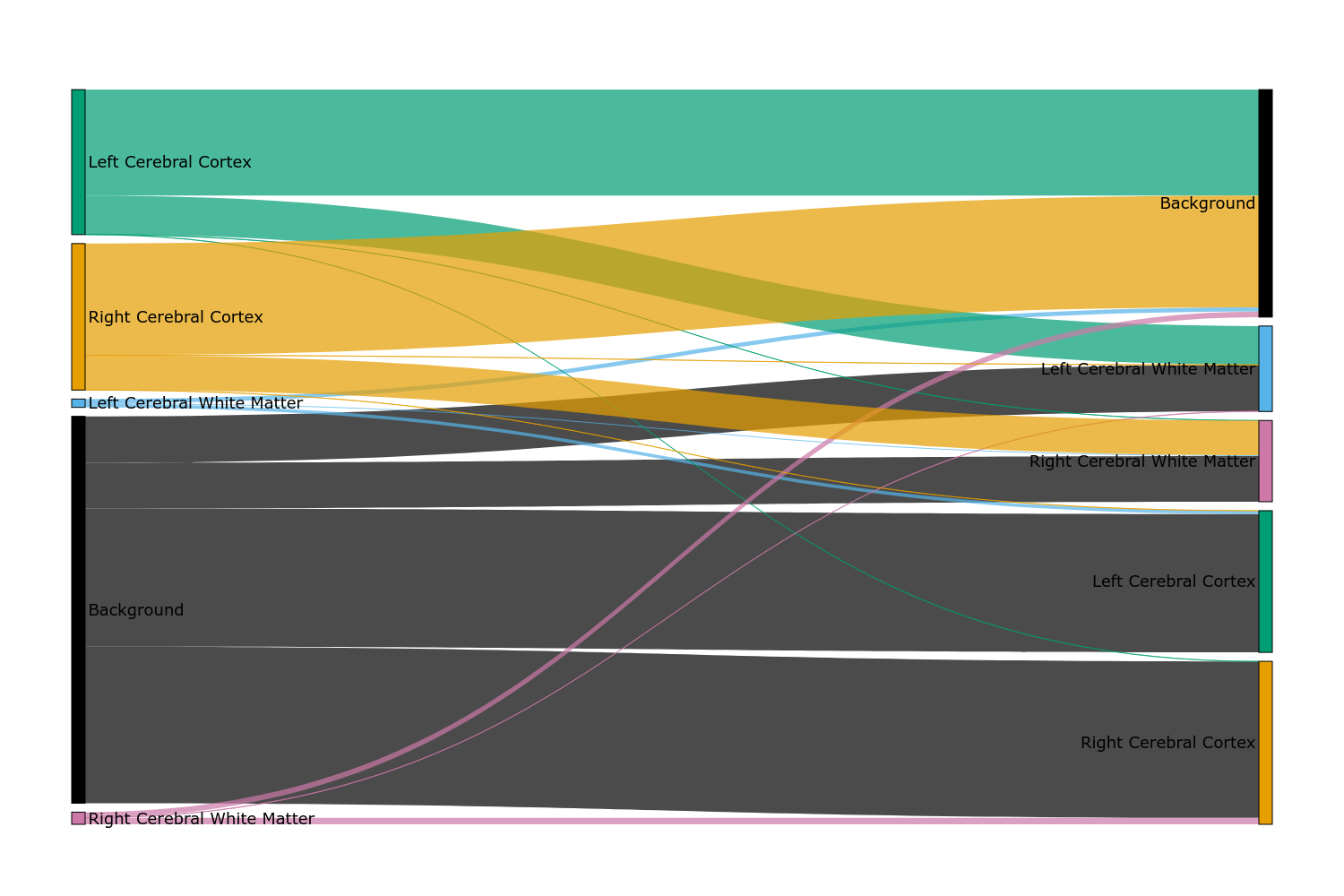}
  \caption{Average change of voxel labels from FreeSurfer (left) to manual segmentation (right) across all subjects. Coloured labels are shown in the sankey diagram as described in Table \ref{tab:labels}.  Cortical labels and background show the largest differences.}
  \label{fig:sankey}
\end{figure}

At a high level, the segmentation procedure can be described as follows: 
\begin{inlinelist}[mode=unboxed]
  \item FreeSurfer generation of surrogate segmentations, 
  \item iterative manual corrections interleaved with expert neuroradiologist review, and  
  \item final validation of the individual segmentations by two neuroradiologists.
\end{inlinelist}
Given the distinctive characteristics and appearance of \qty{9.4}{\tesla} images, we sought to engage the expertise of neuroradiologists with extensive experience in interpreting UHF~data acquired at \qty{9.4}{\tesla}. 
Figure~\ref{fig:mri_slices_with_labels} illustrates an examplar of the segmentation, while Figure~\ref{fig:sankey} shows the average necessity for modifying voxel labels.

\subsubsection*{Initial Segmentation Using FreeSurfer}
In order to optimize the efficiency of the segmentation process, we used FreeSurfer's v7.1.1 \texttt{recon-all -hires} pipeline\cite{zaretskaya_advantages_2018} as a preliminary step.
However, this approach required adaptations due to the fact that its original design did not account for the high field strengths of \qty{9.4}{\tesla} and sub-millimeter resolutions $< \qty{0.8}{\mm}$.
The initial phase of the \texttt{recon-all} pipeline, \texttt{-autorecon1}, includes extensive preprocessing steps such as skull stripping, motion correction, intensity normalization, and Talairach transformation computation.
Following this phase, the generated brain mask was carefully examined.
Should it be necessary, adjustments were made to the pre-flooding height of the watershed algorithm, which is employed to determine a boundary between the brain and the skull.
The skull stripping process was repeated until the desired results were achieved.
For minor corrections, particularly in the dura mater, the convenient \texttt{-gcut} flag was utilized to remove small portions of dura, followed by manual verification of the segmenters.

After these modifications, the manually corrected brain masks were used as input to complete the remaining stages of the \texttt{recon-all} pipeline, \texttt{-autorecon2 -autorecon3}, using the \texttt{-hires} flag.
The result was a detailed cortical ribbon mask delineating the segmented volumes of left and right gray and white matter.
This ribbon, converted to NIfTI format, served as the basis for the subsequent manual segmentation phase.

\subsubsection*{Expert Validation}
The manual segmentation procedure commenced with an initial training of the segmenters.
Following successful completion, the iterative labeling was performed using {ITK-Snap}\cite{yushkevich_user-guided_2006}, allowing for voxel-by-voxel corrections in all three anatomical planes.
Experienced neuroradiologists and neuroscientists performed regular assessments to verify the accuracy of the tissue delineations.
These reviews also served as guidance for the segmenters, providing feedback on the quality of their segmentations and highlighting areas for improvement.
Special attention was given to training the segmenters to accurately determine the boundaries between different tissue classes.
The iterative process continued until the segmentations were deemed satisfactory by the expert reviewers.

Finally, after the manual corrections were completed and approved in a routine review, the segmentations underwent a final validation phase.
This phase involved a thorough inspection in the axial, sagittal, and coronal planes to ensure segmentation consistency, anatomical accuracy, and clear delineation of tissue boundaries.
The final validation was conducted by two expert neuroradiologists, who independently reviewed the segmentations.
Only those segmentations that successfully passed these stringent validations were considered suitable for inclusion in this publication.
This rigorous validation process ensured the high quality and reliability of the segmentations, consistent with the standards required for scientific research in neuroimaging.

\phantomsection\addcontentsline{toc}{section}{Data Records}
\section*{Data Records}

\subsection*{Repository and Dataset Format}
The dataset described here is permanently accessible to the public via OpenNeuro\cite{openneuro} under the accesssion number \href{https://openneuro.org/datasets/ds005216/versions/1.0.1}{ds005216} or via the project page \href{https://ultracortex.org}{https://ultracortex.org}.
The dataset is released under the Creative Commons license~\href{https://creativecommons.org/publicdomain/zero/1.0/}{CC0}, which allows unrestricted reuse of dataset, including for commercial purposes.
The data used in the study were organized using the Brain Imaging Data Structure~(BIDS)\cite{gorgolewski_brain_2016}.
Consequently, the associated metadata are stored in files with tab-separated values~(TSV) as well as with JavaScript Object Notation~(JSON), which include image-specific information about the corresponding demographics and measurement parameters.
For each subject we provide the anonymized $T_1$-weighted image in NIfTI format.
Furthermore, we provide the SynthStrip-generated brain mask and skull-stripped image, the FreeSurfer-produced ribbon segmentation that served as the basis for manual segmentation, and the manual segmentation mask as derivative data.
The detailed data file directory structure is shown in Figure~\ref{fig:data_structure}.

\phantomsection\addcontentsline{toc}{section}{Technical Validation}
\section*{Technical Validation}
\subsection*{Artifacts}
The process of evaluating artifacts involved a meticulous, two-tier inspection methodology by MRI experts.
Initially, after the acquisition, a qualitative control was conducted, examining the images for any prominent artifacts.
\num{88}~images that passed this preliminary inspection were placed into a selection pool.
Subsequently, each image within this pool underwent a thorough visual re-inspection, receiving a binary rating based on its quality.

Images containing significant artifacts, such as those resulting from periodic involuntary motion, sudden involuntary movements, conscious body movements, or hardware and sequence-related issues, were excluded from the selection.
This process yielded a final dataset of \num{86}~images, with the remaining \num{2}~images excluded due to the presence of significant artifacts.

\begin{figure}[ht]
    \centering
    \begin{subfigure}[b]{0.49\textwidth}
    \includegraphics[width=\textwidth]{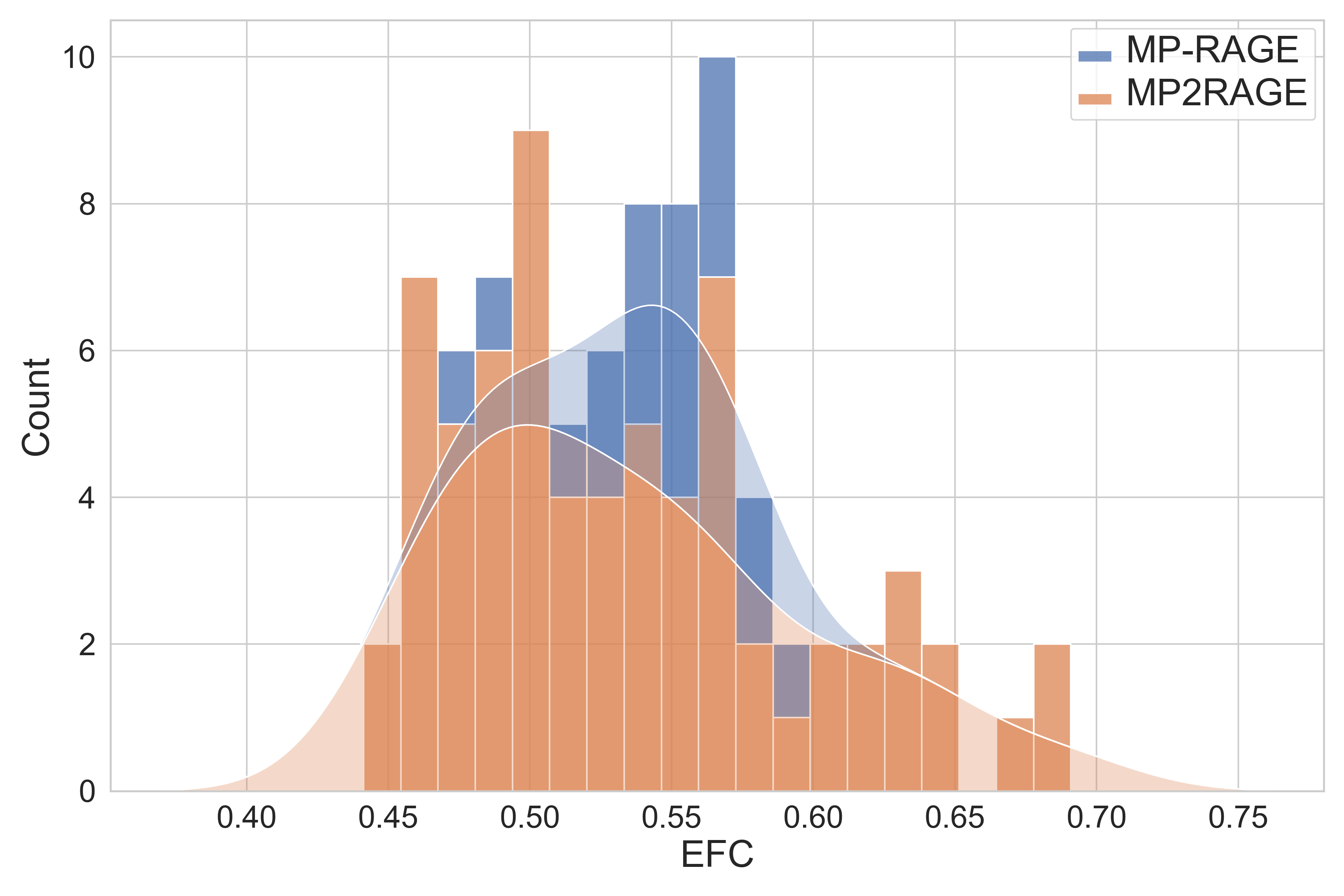}
    \caption{Histogram of Entropy Focus Criterion (EFC).}
    \label{fig:efc}
    \end{subfigure}
    \hfill 
    \begin{subfigure}[b]{0.49\textwidth}
    \includegraphics[width=\textwidth]{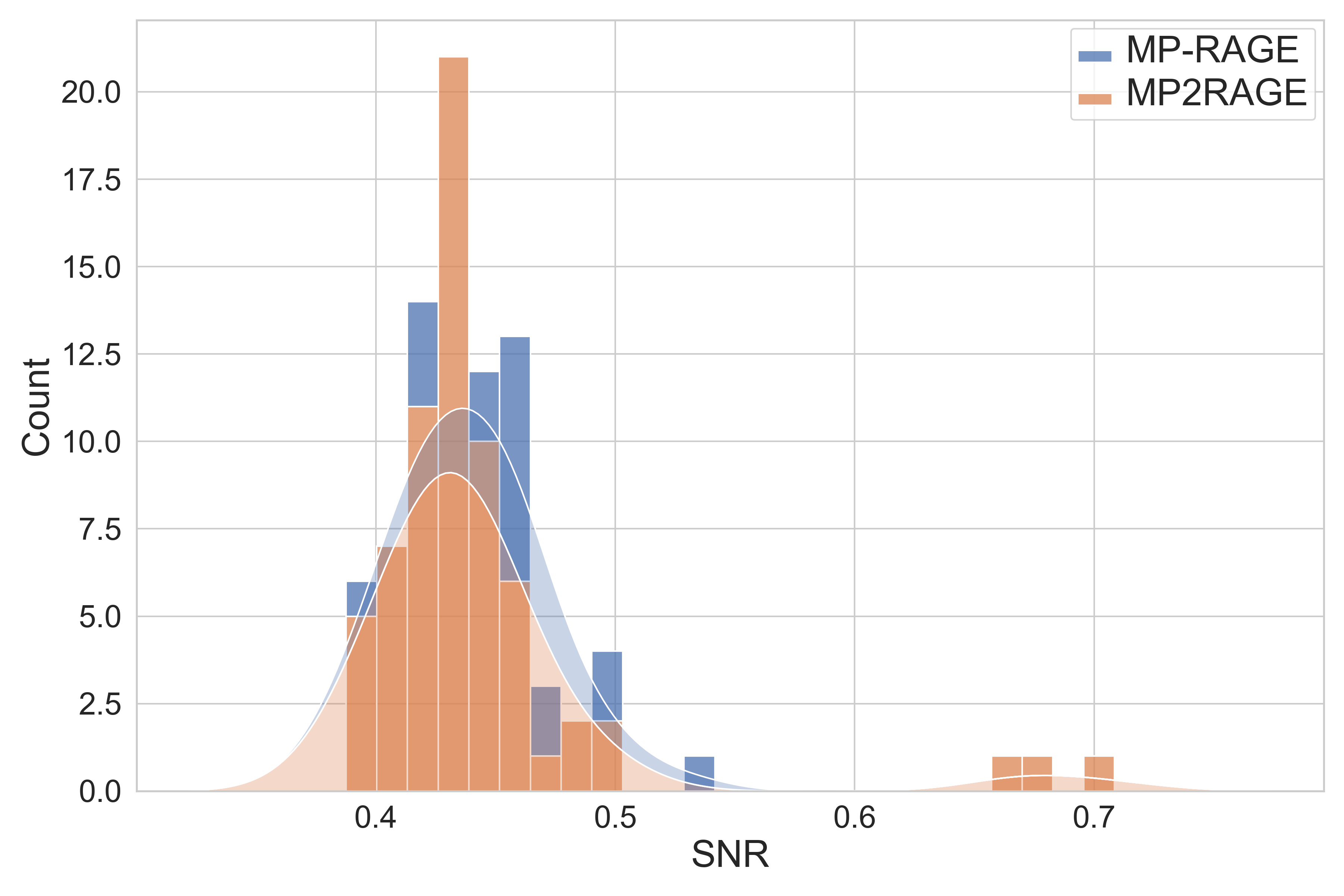}
    \caption{Histogram of Signal-to-Noise Ratio (SNR).}
    \label{fig:snr}
    \end{subfigure}
    \caption{Stacked histograms of image quality metrics with kernel density estimates for MP-RAGE and MP2RAGE sequences.}
    \label{fig:quality_metrics}
\end{figure}

To quantitatively assess the severity of motion artifacts, we calculated the Entropy Focus Criterion $\text{EFC}$, which leverages the Shannon entropy $\text{H}$ of voxel intensities as an indicator of ghosting and blurring due to head motion.
The Shannon entropy of an image with $N$ voxels is defined by the intensity value $x_i$ of voxel $i$ by
\begin{equation}
  \begin{aligned}
    & \text{H} = - \sum_{i=1}^{N} \frac{x_i}{x_\text{max}} \log \left( \frac{x_i}{x_\text{max}} \right) \; \text{, where} \\
    & x_\text{max} = \sqrt{\sum_{i=1}^{N} x_i^2}
  \end{aligned}
\end{equation}
represents the distribution of voxel \enquote{brightness}\cite{atkinson_automatic_1997}.
If all the image energy were concentrated in a single voxel, we would have the greatest possible voxel brightness, $x_\text{max}$.
The $\text{EFC}$ is defined in accordance with MRIQC\cite{esteban_mriqc_2017} as $\text{H}$ normalized by the maximum entropy, which is the case if all voxels have the same value.
This implies that $x_i / x_\text{max} = 1 / \sqrt{N}$, resulting in
\begin{equation}
    \text{EFC} = \left(\sqrt{N} \log \sqrt{N}\right) \cdot \text{H} \; .
\end{equation}
It is preferableto have lower $\text{EFC}$ values, with $\text{EFC} = 0$ indicating maximal energy concentration in a single voxel.
The mean $\text{EFC}$ over all subjects and given sequence configurations is \num[uncertainty-mode = separate]{0.537 \pm 0.057} and a corresponding histogram including kernel density estimate is shown in Figure~\ref{fig:efc}.

\subsection*{Signal-to-Noise Ratio}
To approximate the SNR within the tissue mask, we employed SynthStrip for brain mask extraction.
It is important to note that SynthStrip was not initially designed for UHF~MRI, particularly not for \qty{9.4}{\tesla}.
Consequently, the resulting skull-stripped image is only an approximate representation $\hat{m}$ of the actual brain mask $m$.
The simplified $\text{SNR}_{\hat{m}}$ within the approximated brain mask $\hat{m}$ was calculated in accordance with MRIQC by
\begin{equation}
  \text{SNR}_{\hat{m}} = \frac{\mu_{\hat{m}}}{\sigma_{\hat{m}}} \sqrt{\frac{n_{\hat{m}} - 1}{n_{\hat{m}}}} \; ,
\end{equation}
where $\mu_{\hat{m}}$ is the mean intensity of voxels within the brain mask, $\sigma_{\hat{m}}$ is their standard deviation, and $n_{\hat{m}}$ is the total number of voxels in the brain mask.
The mean $\text{SNR}_{\hat{m}}$ over all subjects is \num[uncertainty-mode = separate]{0.446 \pm 0.053}. 
Figure~\ref{fig:snr} shows a histogram of $\text{SNR}_{\hat{m}}$ for specific sequence configurations.

\subsection*{Segmentations}
For the \num{12} volumes with ground truth segmentations, we conducted additional quantitative evaluations of image quality.
Two key metrics were utilized: the Coefficient of Joint Variation $\text{CJV}$ and the Contrast to Noise Ratio $\text{CNR}$.
The $\text{CJV}$ offers insight into the dispersion of tissue intensities relative to their mean intensity and is expressed as follows:
\begin{equation}
    \text{CJV} = \frac{\sigma_{\text{WM}} + \sigma_{\text{GM}}}{\left| \mu_{\text{WM}} - \mu_{\text{GM}} \right|}
\end{equation}
where $\sigma_{\lbrace \text{WM}, \text{GM} \rbrace}$ and $\mu_{\lbrace \text{WM}, \text{GM} \rbrace}$ represent the standard deviation and the mean intensity of all GM and WM voxels, respectively.
A lower $\text{CJV}$ value indicates a clearer differentiation between tissue classes, with minimal overlap in their intensity distributions.

Furthermore, the $\text{CNR}$ metric was employed to assess the difference in intensity between GM and WM regions relative to the background noise.
A higher $\text{CNR}$ indicates better differentiation between tissue classes.
$\text{CNR}$ is calculated as:
\begin{equation}
    \text{CNR} = \frac{\left|\mu_{\text{GM}} - \mu_{\text{WM}}\right|}{\sqrt{\sigma^2_\text{B} + \sigma_{\text{WM}}^2 + \sigma_{\text{GM}}^2}} \; .
\end{equation}
Here, $\sigma_\text{B}$ denotes the standard deviation of all voxels outside the gray and white matter masks.

The $\text{CNR}$ was determinded to be \num[uncertainty-mode = separate]{0.879 \pm 0.390}, while the $\text{CJV}$ was \num[uncertainty-mode = separate]{1.683 \pm 1.824}. 
These metrics, $\text{CJV}$ and $\text{CNR}$, collectively provide a comprehensive assessment of the segmentation quality in terms of tissue contrast and noise characteristics, thereby ensuring the reliability of the segmentation process in our high-resolution \qty{9.4}{\tesla} MRI dataset.

\phantomsection\addcontentsline{toc}{section}{Usage Notes}
\section*{Usage Notes}
In this study, we have curated a dataset optimized for automated processing to ensure compatibility and ease of integration into various analytical workflows.
The data are provided in widely accepted and well-documented standard formats, including NIfTI and plain text files.
This choice of formats facilitates seamless integration into various analytical environments and eliminates reliance on proprietary software.
In addition, all data processing for this publication was performed using open-source software on conventional computer workstations.
This further enhances the accessibility and reproducibility of our methods.

\phantomsection\addcontentsline{toc}{section}{Code Availability}
\section*{Code Availability}
All processing pipeline scripts are openly available for download.
The code used to generate processed outputs can be accessed via GitHub at \href{https://github.com/MPI-Neuroinformatics/ultracortex}{https://github.com/MPI-Neuroinformatics/ultracortex}.
Further, open-source software packages have been used:
dcm2niix multiple versions v1.0.<X> (\href{https://github.com/rordenlab/dcm2niix}{https://github.com/rordenlab/dcm2niix}),
FreeSurfer v7.1.1 (\href{https://surfer.nmr.mgh.harvard.edu}{https://surfer.nmr.mgh.harvard.edu}),
ITK-Snap v3.8.0 (\href{http://www.itksnap.org}{http://itksnap.org}),
Quickshear v1.1.0 (\href{https://github.com/nipy/quickshear}{https://github.com/nipy/quickshear}), and
SynthStrip as provided by FreeSurfer v7.1.1 (\href{https://surfer.nmr.mgh.harvard.edu/docs/synthstrip}{https://surfer.nmr.mgh.harvard.edu/docs/synthstrip}).

\bibliography{ultracortex_nature_scientific_data}

\begin{thebibliography}{10}
\urlstyle{rm}
\expandafter\ifx\csname url\endcsname\relax
  \def\url#1{\texttt{#1}}\fi
\expandafter\ifx\csname urlprefix\endcsname\relax\def\urlprefix{URL }\fi
\expandafter\ifx\csname doiprefix\endcsname\relax\def\doiprefix{DOI: }\fi
\providecommand{\bibinfo}[2]{#2}
\providecommand{\eprint}[2][]{\url{#2}}

\bibitem{forstmann_multi-modal_2014}
\bibinfo{author}{Forstmann, B.~U.} \emph{et~al.}
\newblock \bibinfo{journal}{\bibinfo{title}{Multi-modal ultra-high resolution
  structural 7-{Tesla} {MRI} data repository}}.
\newblock {\emph{\JournalTitle{Scientific Data}}} \textbf{\bibinfo{volume}{1}},
  \bibinfo{pages}{140050},
  \doiprefix\url{https://doi.org/10.1038/sdata.2014.50} (\bibinfo{year}{2014}).

\bibitem{alkemade_amsterdam_2020}
\bibinfo{author}{Alkemade, A.} \emph{et~al.}
\newblock \bibinfo{journal}{\bibinfo{title}{The {Amsterdam} {Ultra}-high field
  adult lifespan database ({AHEAD}): {A} freely available multimodal 7 {Tesla}
  submillimeter magnetic resonance imaging database}}.
\newblock {\emph{\JournalTitle{NeuroImage}}} \textbf{\bibinfo{volume}{221}},
  \bibinfo{pages}{117200},
  \doiprefix\url{https://doi.org/10.1016/j.neuroimage.2020.117200}
  (\bibinfo{year}{2020}).

\bibitem{svanera_cerebrum-7t_2020}
\bibinfo{author}{Svanera, M.}, \bibinfo{author}{Benini, S.},
  \bibinfo{author}{Bontempi, D.} \& \bibinfo{author}{Muckli, L.}
\newblock \bibinfo{title}{{CEREBRUM}-{7T}: {Fast} and {Fully}-volumetric
  {Brain} {Segmentation} of 7 {Tesla} {MR} {Volumes} [{Data} set]},
  \doiprefix\url{https://doi.org/10.25493/RF12-09N} (\bibinfo{year}{2020}).

\bibitem{pohmann_signal--noise_2016}
\bibinfo{author}{Pohmann, R.}, \bibinfo{author}{Speck, O.} \&
  \bibinfo{author}{Scheffler, K.}
\newblock \bibinfo{journal}{\bibinfo{title}{Signal-to-noise ratio and {MR}
  tissue parameters in human brain imaging at 3, 7, and 9.4 tesla using current
  receive coil arrays}}.
\newblock {\emph{\JournalTitle{Magnetic Resonance in Medicine}}}
  \textbf{\bibinfo{volume}{75}}, \bibinfo{pages}{801--809},
  \doiprefix\url{https://doi.org/10.1002/mrm.25677} (\bibinfo{year}{2016}).

\bibitem{van_essen_wu-minn_2013}
\bibinfo{author}{Van~Essen, D.~C.} \emph{et~al.}
\newblock \bibinfo{journal}{\bibinfo{title}{The {WU}-{Minn} {Human}
  {Connectome} {Project}: {An} overview}}.
\newblock {\emph{\JournalTitle{NeuroImage}}} \textbf{\bibinfo{volume}{80}},
  \bibinfo{pages}{62--79},
  \doiprefix\url{https://doi.org/10.1016/j.neuroimage.2013.05.041}
  (\bibinfo{year}{2013}).

\bibitem{klein_101_2012}
\bibinfo{author}{Klein, A.} \& \bibinfo{author}{Tourville, J.}
\newblock \bibinfo{journal}{\bibinfo{title}{101 {Labeled} {Brain} {Images} and
  a {Consistent} {Human} {Cortical} {Labeling} {Protocol}}}.
\newblock {\emph{\JournalTitle{Frontiers in Neuroscience}}}
  \textbf{\bibinfo{volume}{6}},
  \doiprefix\url{https://doi.org/10.3389/fnins.2012.00171}
  (\bibinfo{year}{2012}).

\bibitem{kuijf_mr_2024}
\bibinfo{author}{Kuijf, H.~J.} \emph{et~al.}
\newblock \bibinfo{title}{{MR} {Brain} {Segmentation} {Challenge} 2018 {Data}},
  \doiprefix\url{https://doi.org/10.34894/E0U32Q} (\bibinfo{year}{2024}).

\bibitem{charyasz_functional_2023}
\bibinfo{author}{Charyasz, E.} \emph{et~al.}
\newblock \bibinfo{journal}{\bibinfo{title}{Functional mapping of sensorimotor
  activation in the human thalamus at 9.4 {Tesla}}}.
\newblock {\emph{\JournalTitle{Frontiers in Neuroscience}}}
  \textbf{\bibinfo{volume}{17}},
  \doiprefix\url{https://doi.org/10.3389/fnins.2023.1116002}
  (\bibinfo{year}{2023}).

\bibitem{shajan_16-channel_2014}
\bibinfo{author}{Shajan, G.} \emph{et~al.}
\newblock \bibinfo{journal}{\bibinfo{title}{A 16-channel dual-row transmit
  array in combination with a 31-element receive array for human brain imaging
  at 9.4 {T}}}.
\newblock {\emph{\JournalTitle{Magnetic Resonance in Medicine}}}
  \textbf{\bibinfo{volume}{71}}, \bibinfo{pages}{870--879},
  \doiprefix\url{https://doi.org/10.1002/mrm.24726} (\bibinfo{year}{2014}).

\bibitem{mugler_iii_rapid_1991}
\bibinfo{author}{Mugler~III, J.~P.} \& \bibinfo{author}{Brookeman, J.~R.}
\newblock \bibinfo{journal}{\bibinfo{title}{Rapid three-dimensional
  {T1}-weighted {MR} imaging with the {MP}-{RAGE} sequence}}.
\newblock {\emph{\JournalTitle{Journal of Magnetic Resonance Imaging}}}
  \textbf{\bibinfo{volume}{1}}, \bibinfo{pages}{561--567},
  \doiprefix\url{https://doi.org/10.1002/jmri.1880010509}
  (\bibinfo{year}{1991}).

\bibitem{marques_mp2rage_2010}
\bibinfo{author}{Marques, J.~P.} \emph{et~al.}
\newblock \bibinfo{journal}{\bibinfo{title}{{MP2RAGE}, a self bias-field
  corrected sequence for improved segmentation and {T1}-mapping at high
  field}}.
\newblock {\emph{\JournalTitle{NeuroImage}}} \textbf{\bibinfo{volume}{49}},
  \bibinfo{pages}{1271--1281},
  \doiprefix\url{https://doi.org/10.1016/j.neuroimage.2009.10.002}
  (\bibinfo{year}{2010}).

\bibitem{hagberg_whole_2017}
\bibinfo{author}{Hagberg, G.} \emph{et~al.}
\newblock \bibinfo{journal}{\bibinfo{title}{Whole brain {MP2RAGE}-based mapping
  of the longitudinal relaxation time at 9.{4T}}}.
\newblock {\emph{\JournalTitle{NeuroImage}}} \textbf{\bibinfo{volume}{144}},
  \bibinfo{pages}{203--216},
  \doiprefix\url{https://doi.org/10.1016/j.neuroimage.2016.09.047}
  (\bibinfo{year}{2017}).

\bibitem{hurley_tailored_2010}
\bibinfo{author}{Hurley, A.~C.} \emph{et~al.}
\newblock \bibinfo{journal}{\bibinfo{title}{Tailored {RF} pulse for
  magnetization inversion at ultrahigh field}}.
\newblock {\emph{\JournalTitle{Magnetic Resonance in Medicine}}}
  \textbf{\bibinfo{volume}{63}}, \bibinfo{pages}{51--58},
  \doiprefix\url{10.1002/mrm.22167} (\bibinfo{year}{2010}).
\newblock \bibinfo{note}{\_eprint:
  https://onlinelibrary.wiley.com/doi/pdf/10.1002/mrm.22167}.

\bibitem{griswold_generalized_2002}
\bibinfo{author}{Griswold, M.~A.} \emph{et~al.}
\newblock \bibinfo{journal}{\bibinfo{title}{Generalized autocalibrating
  partially parallel acquisitions ({GRAPPA})}}.
\newblock {\emph{\JournalTitle{Magnetic Resonance in Medicine}}}
  \textbf{\bibinfo{volume}{47}}, \bibinfo{pages}{1202--1210},
  \doiprefix\url{https://doi.org/10.1002/mrm.10171} (\bibinfo{year}{2002}).

\bibitem{breuer_controlled_2005}
\bibinfo{author}{Breuer, F.~A.} \emph{et~al.}
\newblock \bibinfo{journal}{\bibinfo{title}{Controlled aliasing in parallel
  imaging results in higher acceleration ({CAIPIRINHA}) for multi-slice
  imaging}}.
\newblock {\emph{\JournalTitle{Magnetic Resonance in Medicine}}}
  \textbf{\bibinfo{volume}{53}}, \bibinfo{pages}{684--691},
  \doiprefix\url{https://doi.org/10.1002/mrm.20401} (\bibinfo{year}{2005}).

\bibitem{obrien_robust_2014}
\bibinfo{author}{O'Brien, K.~R.} \emph{et~al.}
\newblock \bibinfo{journal}{\bibinfo{title}{Robust {T1}-{Weighted} {Structural}
  {Brain} {Imaging} and {Morphometry} at {7T} {Using} {MP2RAGE}}}.
\newblock {\emph{\JournalTitle{PLOS ONE}}} \textbf{\bibinfo{volume}{9}},
  \bibinfo{pages}{e99676},
  \doiprefix\url{https://doi.org/10.1371/journal.pone.0099676}
  (\bibinfo{year}{2014}).

\bibitem{li_first_2016}
\bibinfo{author}{Li, X.}, \bibinfo{author}{Morgan, P.~S.},
  \bibinfo{author}{Ashburner, J.}, \bibinfo{author}{Smith, J.} \&
  \bibinfo{author}{Rorden, C.}
\newblock \bibinfo{journal}{\bibinfo{title}{The first step for neuroimaging
  data analysis: {DICOM} to {NIfTI} conversion}}.
\newblock {\emph{\JournalTitle{Journal of Neuroscience Methods}}}
  \textbf{\bibinfo{volume}{264}}, \bibinfo{pages}{47--56},
  \doiprefix\url{https://doi.org/10.1016/j.jneumeth.2016.03.001}
  (\bibinfo{year}{2016}).

\bibitem{hoopes_synthstrip_2022}
\bibinfo{author}{Hoopes, A.}, \bibinfo{author}{Mora, J.~S.},
  \bibinfo{author}{Dalca, A.~V.}, \bibinfo{author}{Fischl, B.} \&
  \bibinfo{author}{Hoffmann, M.}
\newblock \bibinfo{journal}{\bibinfo{title}{{SynthStrip}: skull-stripping for
  any brain image}}.
\newblock {\emph{\JournalTitle{NeuroImage}}} \textbf{\bibinfo{volume}{260}},
  \bibinfo{pages}{119474},
  \doiprefix\url{https://doi.org/10.1016/j.neuroimage.2022.119474}
  (\bibinfo{year}{2022}).

\bibitem{schimke_quickshear_2011}
\bibinfo{author}{Schimke, N.} \& \bibinfo{author}{Hale, J.}
\newblock \bibinfo{title}{Quickshear {Defacing} for {Neuroimages}}.
\newblock In \emph{\bibinfo{booktitle}{Proceedings of the 2nd {USENIX}
  conference on {Health} security and privacy}}, {HealthSec}'11,
  \bibinfo{pages}{11} (\bibinfo{publisher}{USENIX Association},
  \bibinfo{address}{USA}, \bibinfo{year}{2011}).

\bibitem{fischl_whole_2002}
\bibinfo{author}{Fischl, B.} \emph{et~al.}
\newblock \bibinfo{journal}{\bibinfo{title}{Whole {Brain} {Segmentation}:
  {Automated} {Labeling} of {Neuroanatomical} {Structures} in the {Human}
  {Brain}}}.
\newblock {\emph{\JournalTitle{Neuron}}} \textbf{\bibinfo{volume}{33}},
  \bibinfo{pages}{341--355},
  \doiprefix\url{https://doi.org/10.1016/S0896-6273(02)00569-X}
  (\bibinfo{year}{2002}).

\bibitem{zaretskaya_advantages_2018}
\bibinfo{author}{Zaretskaya, N.}, \bibinfo{author}{Fischl, B.},
  \bibinfo{author}{Reuter, M.}, \bibinfo{author}{Renvall, V.} \&
  \bibinfo{author}{Polimeni, J.~R.}
\newblock \bibinfo{journal}{\bibinfo{title}{Advantages of cortical surface
  reconstruction using submillimeter {7} {T} {MEMPRAGE}}}.
\newblock {\emph{\JournalTitle{NeuroImage}}} \textbf{\bibinfo{volume}{165}},
  \bibinfo{pages}{11--26},
  \doiprefix\url{https://doi.org/10.1016/j.neuroimage.2017.09.060}
  (\bibinfo{year}{2018}).

\bibitem{yushkevich_user-guided_2006}
\bibinfo{author}{Yushkevich, P.~A.} \emph{et~al.}
\newblock \bibinfo{journal}{\bibinfo{title}{User-guided {3D} active contour
  segmentation of anatomical structures: {Significantly} improved efficiency
  and reliability}}.
\newblock {\emph{\JournalTitle{NeuroImage}}} \textbf{\bibinfo{volume}{31}},
  \bibinfo{pages}{1116--1128},
  \doiprefix\url{https://doi.org/10.1016/j.neuroimage.2006.01.015}
  (\bibinfo{year}{2006}).

\bibitem{openneuro}
\bibinfo{author}{Mahler, L.} \emph{et~al.}
\newblock \bibinfo{title}{"ultracortex: Submillimeter ultra-high field 9.4t
  brain mr image collection and manual cortical segmentations"},
  \doiprefix\url{doi:10.18112/openneuro.ds005216.v1.0.1}
  (\bibinfo{year}{2024}).

\bibitem{gorgolewski_brain_2016}
\bibinfo{author}{Gorgolewski, K.~J.} \emph{et~al.}
\newblock \bibinfo{journal}{\bibinfo{title}{The brain imaging data structure, a
  format for organizing and describing outputs of neuroimaging experiments}}.
\newblock {\emph{\JournalTitle{Scientific Data}}} \textbf{\bibinfo{volume}{3}},
  \bibinfo{pages}{160044},
  \doiprefix\url{https://doi.org/10.1038/sdata.2016.44} (\bibinfo{year}{2016}).

\bibitem{atkinson_automatic_1997}
\bibinfo{author}{Atkinson, D.}, \bibinfo{author}{Hill, D.},
  \bibinfo{author}{Stoyle, P.}, \bibinfo{author}{Summers, P.} \&
  \bibinfo{author}{Keevil, S.}
\newblock \bibinfo{journal}{\bibinfo{title}{Automatic correction of motion
  artifacts in magnetic resonance images using an entropy focus criterion}}.
\newblock {\emph{\JournalTitle{IEEE Transactions on Medical Imaging}}}
  \textbf{\bibinfo{volume}{16}}, \bibinfo{pages}{903--910},
  \doiprefix\url{https://doi.org/10.1109/42.650886} (\bibinfo{year}{1997}).

\bibitem{esteban_mriqc_2017}
\bibinfo{author}{Esteban, O.} \emph{et~al.}
\newblock \bibinfo{journal}{\bibinfo{title}{{MRIQC}: {Advancing} the automatic
  prediction of image quality in {MRI} from unseen sites}}.
\newblock {\emph{\JournalTitle{PLOS ONE}}} \textbf{\bibinfo{volume}{12}},
  \bibinfo{pages}{e0184661},
  \doiprefix\url{https://doi.org/10.1371/journal.pone.0184661}
  (\bibinfo{year}{2017}).

\end{thebibliography}

\phantomsection\addcontentsline{toc}{section}{Acknowledgements}
\section*{Acknowledgements}
We would like to express our sincere gratitude to all student assistants who played an invaluable role in creating the manual segmentations for this project.
Their meticulous efforts, dedication, and attention to detail were critical to the successful completion of this task.
We deeply appreciate their hard work and dedication, which greatly enhanced the overall quality and reliability of our research.
We gratefully acknowledge funding by the Max Planck Society.
This work was supported by the German Research Council (DFG) Grant GZ: GR 833/11-1.  

\phantomsection\addcontentsline{toc}{section}{Author Contributions Statement}
\section*{Author Contributions Statement}
Participant recruitment, data acquisition, and qualitative data control: J.B., F.B., E.C., M.E., G.H., R.H., V.K., P.M., and D.R.;
Conception, data collection, qualitative data control, and cleaning: J.S.;
Design: D.R., L.M., and J.S.;
Data processing and anonymization: L.M. and J.S.;
Technical validation: L.M.;
Supervision of the segmentation process: D.R. and L.M.;
Feedback to segmentors and final validation of segmentation: B.B. and T.L.;
Writing of the manuscript: L.M. and J.S.;
Supervision, participation in discussions and suggestions for manuscript improvement: B.B., G.L., and T.L.;
Resources, supervision, and funding acquisition: K.S.; 
All authors contributed to the article and approved the submitted version.

\phantomsection\addcontentsline{toc}{section}{Competing Interests}
\section*{Competing Interests}
J.S. -- employee of AIRAmed GmbH, unrelated to the content of this manuscript.
B.B. -- co-founder, shareholder, and CTO of AIRAmed GmbH, unrelated to the content of this manuscript.
T.L. -- co-founder, shareholder, and CEO of AIRAmed GmbH, unrelated to the content of this manuscript.
P.M. -- has received honoraria as an advisory board member from Biogen and Alexion, unrelated to the content of this manuscript.
The other authors declare no competing interests.

\phantomsection\addcontentsline{toc}{section}{Figures \& Tables}
\section*{Figures \& Tables}

\begin{table}[ht]
  \centering
  \begin{tabular}{l r r r r}
    \toprule
    Age Group \lbrack \unit{\years}\rbrack & Images \lbrack n (\unit{\percent})\rbrack & Median Age \lbrack \unit{years} (IQR)\rbrack & Males \lbrack n\rbrack & Females \lbrack n\rbrack \\
    \midrule
    \numrange[range-phrase = {\,--\,}]{20}{25} & \num{16}\,(\num{19}) & \num{22}\;(\numrange[range-phrase = {\,--\,}]{22}{23}) & \num{10} & \num{6} \\
    \numrange[range-phrase = {\,--\,}]{25}{30} & \num{35}\,(\num{42}) & \num{28}\;(\numrange[range-phrase = {\,--\,}]{26}{29}) & \num{24} & \num{11} \\
    \numrange[range-phrase = {\,--\,}]{30}{35} & \num{17}\,(\num{20}) & \num{31}\;(\numrange[range-phrase = {\,--\,}]{30}{32}) & \num{14} & \num{3} \\
    \numrange[range-phrase = {\,--\,}]{35}{40} & \num{7}\,(\num{8}) & \num{35}\;(\numrange[range-phrase = {\,--\,}]{35}{37}) & \num{3} & \num{4} \\
    \numrange[range-phrase = {\,--\,}]{40}{45} & \num{7}\,(\num{8}) & \num{42}\;(\numrange[range-phrase = {\,--\,}]{41}{43}) & \num{2} & \num{5} \\
    \numrange[range-phrase = {\,--\,}]{45}{50} & \num{3}\,(\num{4}) & \num{49}\;(\numrange[range-phrase = {\,--\,}]{47}{49}) & \num{0} & \num{3} \\
    \numrange[range-phrase = {\,--\,}]{50}{55} & \num{1}\,(\num{1}) & \num{53}\;(\numrange[range-phrase = {\,--\,}]{53}{53}) & \num{1} & \num{0} \\
    \bottomrule
  \end{tabular}
  \caption{\label{tab:demographics}Age and sex characteristics of dataset.}
\end{table}

\begin{table}[ht]
  \centering
  \begin{tabular}{l r r r r r r r r r}
    \toprule
    Sequence & Images & Voxel Size \lbrack\unit{\mm^3}\rbrack & slices & matrix & TR \lbrack\unit{\milli\second}\rbrack & TE \lbrack\unit{\milli\second}\rbrack & FA \lbrack\unit{\degree}\rbrack & R & CAIPI \\
    \midrule
    MP-RAGE & 18 \\
        & 4 & \numproduct{0.7 x 0.7 x 0.7} & 240 & \numproduct{300 x 306} & \num{6500} & \num{3.4} & \num{7.5} & \numproduct{2 x 1} & off \\
        & 1 & \numproduct{0.7 x 0.7 x 0.7} & 256 & \numproduct{300 x 306} & \num{6500} & \num{3.4} & \num{7.5} & \numproduct{2 x 1} & off \\
        & 1 & \numproduct{0.6 x 0.6 x 0.6} & 288 & \numproduct{352 x 352} & \num{6000} & \num{2.3} & \num{8.0} & \numproduct{2 x 2} & on \\
        & 4 & \numproduct{0.6 x 0.6 x 0.6} & 288 & \numproduct{352 x 352} & \num{3800} & \num{2.5} & \num{6.0} & \numproduct{2 x 2} & on \\
        & 8 & \numproduct{0.6 x 0.6 x 0.6} & 288 & \numproduct{320 x 320} & \num{3800} & \num{2.5} & \num{6.0} & \numproduct{2 x 2} & on \\
    \\
    MP2RAGE & 68 \\
        &11 & \numproduct{0.8 x 0.8 x 0.8} & 192 & \numproduct{256 x 256} & \num{9000} & \num{2.3} & \num{4.0} & \numproduct{3 x 1} & off \\
        &20 & \numproduct{0.8 x 0.8 x 0.8} & 236 & \numproduct{256 x 256} & \num{6000} & \num{2.3} & \num{4.0} & \numproduct{3 x 1} & off \\
        & 1 & \numproduct{0.6 x 0.6 x 0.6} & 256 & \numproduct{352 x 352} & \num{6000} & \num{3.0} & \num{4.0} & \numproduct{2 x 2} & on \\
        & 6 & \numproduct{0.6 x 0.6 x 0.6} & 256 & \numproduct{352 x 352} & \num{6000} & \num{3.0} & \num{5.0} & \numproduct{2 x 2} & off \\
        &27 & \numproduct{0.6 x 0.6 x 0.6} & 256 & \numproduct{352 x 352} & \num{6000} & \num{3.0} & \num{5.0} & \numproduct{2 x 2} & on \\
        & 3 & \numproduct{0.6 x 0.6 x 0.6} & 260 & \numproduct{261 x 315} & \num{6000} & \num{3.0} & \num{5.0} & \numproduct{2 x 2} & on \\
    \bottomrule
  \end{tabular}
  \caption{\label{tab:scanning_parameters}Details of the dataset, outlining the various scanning parameter sets. MR Images acquired at the Department for High-field Magnetic Resonance of the Max Planck Institute for Biological Cybernetics, T\"ubingen, Germany.}
\end{table}
\begin{table}[ht]
  \centering
  \begin{tabular}{l r l l}
    \toprule
    Structure & ID & Color & (R, G, B)\\
    \midrule
    Background & 0 & black/transparent & (0, 0, 0)\\
    Left cerebral white matter & 2 & sky blue &(86, 180, 133) \\
    Left cerebral cortex & 3 & blueish green &(0, 158, 115) \\
    Right cerebral white matter & 41 & reddish purple &(204, 121, 167) \\
    Right cerebral cortex & 42 & orange &(230, 159, 0) \\
    \bottomrule
  \end{tabular}
  \caption{\label{tab:labels}UltraCortex segmentation IDs and mapping to color scheme in Figure~\ref{fig:mri_slices_with_labels}.}
\end{table}

\begin{figure}[ht]
  \dirtree{%
    .1 ./ultracortex.
    .2 dataset\_description.json.
    .2 LICENSE.
    .2 participants.tsv.
    .2 README.md.
    .2 derivatives.
    .3 brainmasks.
    .4 sub-<participant ID>\_ses-<session ID>\_brainmask.nii.
    .3 freesurfer\_segmentation.
    .4 sub-<participant ID>\_ses-<session ID>\_seg.nii.
    .3 manual\_segmentation.
    .4 sub-<participant ID>\_ses-<session ID>\_seg.nii.
    .3 skullstrips.
    .4 sub-<participant ID>\_ses-<session ID>\_skullstrip.nii.
    .3 scanning\_parameters.json.
    .3 scanning\_parameters.tsv.
    .2 sub-<participant ID>.
    .3 ses-<session ID>.
    .4 anat.
    .5 sub-<participant ID>\_ses-<session ID>\_T1w.json.
    .5 sub-<participant ID>\_ses-<session ID>\_T1w.nii.
    .2 sub-\dots.
  }
  \caption{Structure of files and directories.}
  \label{fig:data_structure}
\end{figure}

\end{document}